\documentclass[10pt,twocolumn,letterpaper]{article}

\usepackage[pagenumbers]{cvpr} 
\usepackage{multirow}

\usepackage[dvipsnames]{xcolor}

\definecolor{cvprblue}{rgb}{0.21,0.49,0.74}
\usepackage[pagebackref,breaklinks,colorlinks,citecolor=cvprblue]{hyperref}

\def\paperID{7093} 
\def\confName{CVPR}
\def\confYear{2025}

\title{STAR-Edge: Structure-aware Local Spherical Curve Representation for Thin-walled Edge Extraction from Unstructured Point Clouds}

\author{
    Zikuan Li\textsuperscript{*}, Honghua Chen\textsuperscript{*}, Yuecheng Wang, Sibo Wu, Mingqiang Wei, Jun Wang\textsuperscript{\dag}\\
    Nanjing University of Aeronautics and Astronautics\\
}

\begin{document}
\twocolumn[{%
\renewcommand\twocolumn[1][]{#1}%
\maketitle

\begin{center} 
    \vspace{-5mm}
    \setlength{\abovecaptionskip}{12pt}
    \centering
    \includegraphics[width=1\linewidth]{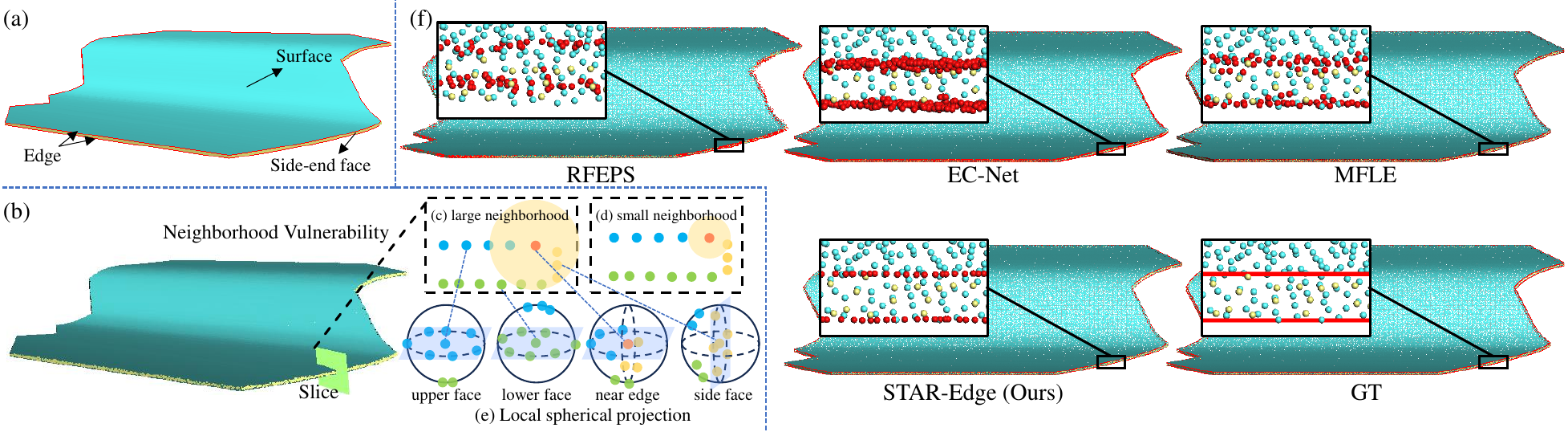}
    \captionof{figure}{The characteristics of our method and visual comparison of edge extraction results in thin-walled structures.
(a) defines the concept of a thin-walled edge.
(b) highlights that the primary challenge in thin-walled edge extraction is the sensitivity to local neighborhood selection.
(c) illustrates how a larger neighborhood may include points from both upper and lower surfaces as well as side-end faces, blurring boundary information.
(d) shows that a smaller neighborhood may suffer from noise and insufficient sampling on side faces, lacking the contextual information needed for accurate edge point recognition.
(e) depicts the spherical projection of the local neighborhood of four points, where points of the same color lie on the same underlying surface. This illustrates our key observation that points co-planar with the neighborhood center tend to align along a great circle arc after spherical projection.
(f) provides a qualitative comparison of our STAR-Edge with RFEPS~\cite{xu2022rfeps}, EC-Net~\cite{yu2018ec}, and MFLE~\cite{chen2021multiscale}.} 
    \label{fig:teaser}
\end{center}

}]

\footnotetext{\textsuperscript{*} Equal contribution \textsuperscript{\dag} Corresponding author}

\begin{abstract}

Extracting geometric edges from unstructured point clouds remains a significant challenge, particularly in thin-walled structures that are commonly found in everyday objects. Traditional geometric methods and recent learning-based approaches frequently struggle with these structures, as both rely heavily on sufficient contextual information from local point neighborhoods. 
However, 3D measurement data of thin-walled structures often lack the accurate, dense, and regular neighborhood sampling required for reliable edge extraction, resulting in degraded performance.

In this work, we introduce STAR-Edge, a novel approach designed for detecting and refining edge points in thin-walled structures. Our method leverages a unique representation—the local spherical curve—to create structure-aware neighborhoods that emphasize co-planar points while reducing interference from close-by, non-co-planar surfaces. 
This representation is transformed into a rotation-invariant descriptor, which, combined with a lightweight multi-layer perceptron, enables robust edge point classification even in the presence of noise and sparse or irregular sampling. Besides, we also use the local spherical curve representation to estimate more precise normals and introduce an optimization function to project initially identified edge points exactly on the true edges.
Experiments conducted on the ABC dataset and thin-walled structure-specific datasets demonstrate that STAR-Edge outperforms existing edge detection methods, showcasing better robustness under various challenging conditions. The source code is available at \url{https://github.com/Miraclelzk/STAR-Edge}.

\end{abstract}    
\section{Introduction}
\label{sec:intro}
Thin-walled structures are essential in our daily lives, seen in items like mechanical parts, furniture, and various manufactured objects.
These structures often contain thin-walled edges, defined as boundary lines where surfaces intersect at their side-end faces, characterized by a minimal thickness relative to other dimensions (see \cref{fig:teaser} (a)).
As geometric edges provide an abstraction of complex 3D shapes, extracting edges from 3D measurements (e.g. unstructured point clouds) in these structures is crucial for downstream tasks. 

Extracting geometric edges from point clouds has been a long-standing problem. 
Traditional techniques generally rely on local geometric properties, such as eigenvalues of the covariance matrix~\cite{xia2017fast, bazazian2015fast}, normals~\cite{weber2010sharp}, curvature~\cite{hackel2016contour}, or elaborately-designed descriptors~\cite{chen2021multiscale}.
While effective to some extent, these methods depend on first- or second-order derivatives estimated from local neighborhoods, making them particularly sensitive to point disturbances and sampling sparsity or irregularities, especially on thick side-end faces.
More recently, some deep neural  networks~\cite{yu2018ec,bazazian2021edc,himeur2021pcednet,liu2021pcwf} have been proposed to enhance edge accuracy by leveraging deep features~\cite{wang2020pie}, neural edge representations~\cite{zhu2023nerve}, or novel point representations (e.g., Scale Space Matrix~\cite{liu2021pcwf}). However, these methods still struggle with thin-walled structures (as seen in \cref{fig:teaser} (f)), as they rely on extracting useful information from \textit{valid} local neighborhoods.

To be more specific, the local neighborhood of points around thin-walled edges is often unreliable. A larger neighborhood may include points from both upper and lower surfaces as well as side-end faces, blurring the boundary information (see \cref{fig:teaser} (c)). 
Conversely, a smaller neighborhood may suffer from noise and insufficient sampling on side-end faces, lacking the contextual information needed for accurate edge point recognition by those above networks (see \cref{fig:teaser} (d)). 
Although multi-scale schemes~\cite{wang2020pie} can partially mitigate this issue, we argue that this approach is sub-optimal, as it requires hard-coding several fixed neighborhood scales.

Ideally, each point's local representation should be \textit{structure-aware}. 
This means it should reliably emphasize points on the same underlying surface (e.g., the same plane) while minimizing interference from points on different structures (e.g., another plane). 
Interestingly, we observe that points co-planar with the neighborhood center tend to align along a great circle arc (dashed circle on the blue plane in \cref{fig:teaser} (e)) after spherical projection, effectively reflecting the underlying surface of the thin-walled structure.
Based on this observation, naturally, we can construct a local spherical curve to incorporate as many points on the great circle arc as possible, thereby enhancing points on the same underlying surface while reducing the influence of points from other surfaces.

To this end, we present STAR-Edge, a novel approach for detecting and refining edge points in thin-walled structures. Our method begins with a new representation, the local spherical curve, to explicitly build structure-aware local neighborhoods. This design enhances information aggregation from nearby co-planar points, even in noisy, sparse, or irregularly point distribution settings.
Next, we construct a rotation-invariant descriptor based on the local spherical curve and integrate it with a multi-layer perceptron (MLP) to accurately classify edge and non-edge points. Beyond edge classification, we also use the local spherical curve representation to estimate more precise point normals and introduce an optimization function to project points closer to the true edge.
To validate our approach, we conduct a series of experiments on the well-known ABC dataset~\cite{koch2019abc} and specialized datasets for thin-walled structures~\cite{chen2024thin}. Extensive experimental results demonstrate that our approach outperforms existing edge detection methods.
Our contributions are summarized as follows:
\begin{itemize}
\item{We introduce STAR-Edge, a new method for detecting and refining edge points in thin-walled structures. }
\item{We develop a new representation, the local spherical curve, to construct structure-aware local neighborhoods, enhancing information aggregation from nearby co-planar points, even under challenging conditions like noise, sparsity, or irregular point distribution.}
\item{Using the local spherical curve representation, we construct a rotation-invariant local descriptor to achieve more accurate edge point classification. Additionally, this representation allows us to accurately calculate point cloud normals, enabling further refinement of edge points.}
\item{We perform extensive experiments on specialized datasets for thin-walled structures and the ABC dataset, showing that our method outperforms existing edge extraction techniques.}
\end{itemize}

\section{Related Works}
\label{sec:Related Works}
\subsection{Traditional method for edge extraction}
Early edge extraction techniques primarily targeted mesh models~\cite{hubeli2001multiresolution, wang2006incremental}, with subsequent advancements focusing on direct edge feature extraction from point clouds.
Oztireli et al.~\cite{oztireli2009feature} apply robust kernel regression combined with Moving Least Squares (MLS) to preserve fine details and sharp features in point-based surface representations. 
Daniels et al.~\cite{daniels2007robust} further advance this by introducing a robust method that identifies sharp features in point clouds by generating smooth edge-aligned curves through multi-step refinement and surface fitting.
They both rely on iterative refinement processes to progressively extract and enhance sharp edge features in point cloud data.

Mérigot et al.~\cite{merigot2010voronoi} propose a robust method using Voronoi-based covariance matrices to robustly extract curvature, sharp features, and normals from point clouds. 
Bazazian et al.~\cite{bazazian2015fast} present a fast method for sharp edge extraction from unorganized point clouds using eigenvalues of covariance matrices from k-nearest neighbors. 
Building on these approaches, other researchers have explored various techniques for identifying and preserving critical features in point clouds. For instance, Huang et al.~\cite{huang2013edge} present an edge-aware resampling method that computes reliable normals away from edges while effectively preserving sharp features. 
Ni et al.~\cite{ni2016edge} propose AGPN, a method for detecting edges and tracing feature lines in 3D point clouds using neighborhood geometry analysis with RANSAC and hybrid region growing. 
Mineo et al.~\cite{mineo2019novel} present a boundary point detection algorithm with fast Fourier transform (FFT)-based filtering to improve edge reconstruction and reduce noise in tessellated surfaces from point clouds.
Chen et al.~\cite{chen2021multiscale} propose a two-phase algorithm for extracting line-type features from unstructured point clouds, using a statistical metric to detect feature points and an anisotropic contracting scheme to reconstruct feature lines against noise. 

In summary, most traditional point cloud edge extraction methods rely heavily on neighborhood reliability, which includes dense sampling and minimal noise. However, in thin-walled structures, neighborhood information is often ambiguous, making conventional methods generally unsuitable for such structures.

\subsection{Learning-based methods for edge extraction}
With the emergence of learning techniques, many data-driven approaches have been proposed to improve the robustness of edge extraction and the preservation of sharp edges~\cite{chen2022repcd,zhou2020geometry}.
EC-Net~\cite{yu2018ec} is introduced as an edge-aware point set consolidation network designed to handle sparse, irregular, and noisy point clouds by both minimizing distances to 3D meshes and edges.
PIE-NET~\cite{wang2020pie} uses PointNet++~\cite{qi2017pointnet++} as the backbone for classification and regression to extract edges from 3D point clouds as parametric curves, and has been extensively validated on synthetic data. 
CPE-Net~\cite{chen2024aircraft} is designed in a multitask learning paradigm and uses a learnable weighted least squares fitting approach to accurately extract countersink edges under noisy conditions.
Generally, all of these methods initially classify edge points using deep learning techniques, followed by edge generation through upsampling, clustering, or fitting smooth curves to capture edge structures in point clouds.

There are also some methods that focus on learning edge features directly from point clouds.
Bode et al.~\cite{bode2023bounded} proposed BoundED, a neural network that uses local neighborhood statistics to classify non-edge, sharp-edge, and boundary points in 3D point clouds. 
Yao et al.~\cite{yao2023hgnet} propose HGNet, a point cloud network that aggregates hierarchical features from points to super-surfaces. 
Zhao et al.~\cite{zhao2023sharp} propose a deep learning approach to detect and consolidate sharp feature edges in 3D point clouds using a multi-task network that predicts displacement vectors for edge refinement.
APEE-Net~\cite{chen2024thin} is a dual-task neural network designed for extracting precise edges on aircraft panels by identifying nearby points and also using displacement vectors for repositioning.
In addition, NerVE~\cite{zhu2023nerve} defines a neural volumetric edge representation to extract parametric edge curves through graph search, leveraging neural fields rather than traditional point cloud processing. 
Ye et al.~\cite{ye2023nef} propose a Neural Edge Field method for reconstructing 3D feature curves from multi-view images, optimizing a neural implicit field with a rendering loss to extract 3D edges without 3D supervision. 
Although these learning-based methods achieve better performance than previous geometric-based methods, they still rely heavily on sufficient contextual information from local point neighborhoods, making them challenging to apply to thin-walled structures.

\section{Method}
\label{sec:Method}
\begin{figure*}[t]
    \centering
    \includegraphics[width=0.9\linewidth]{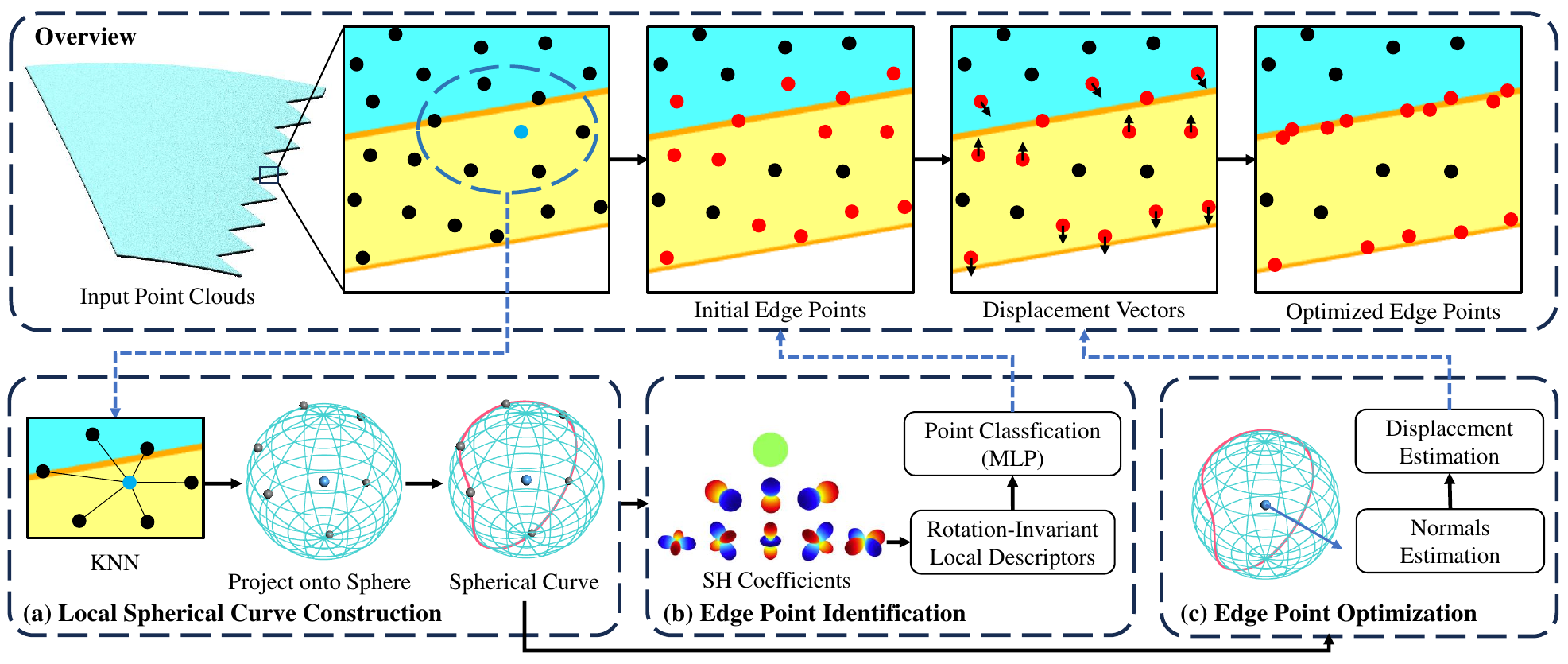}
    \caption{Overview of the proposed method. STAR-edge comprises three main steps: 
    (a) constructing a local spherical curve;
    (b) identifying edge points by integrating rotation-invariant local descriptors with a light-weight MLP layer; 
    and (c) optimizing edge points by adjusting their positions based on the estimated normal vector from the local spherical curve.}
    \label{fig:overview}
\end{figure*}

\begin{figure*}[t]
\centering
\includegraphics[width=0.9\linewidth]{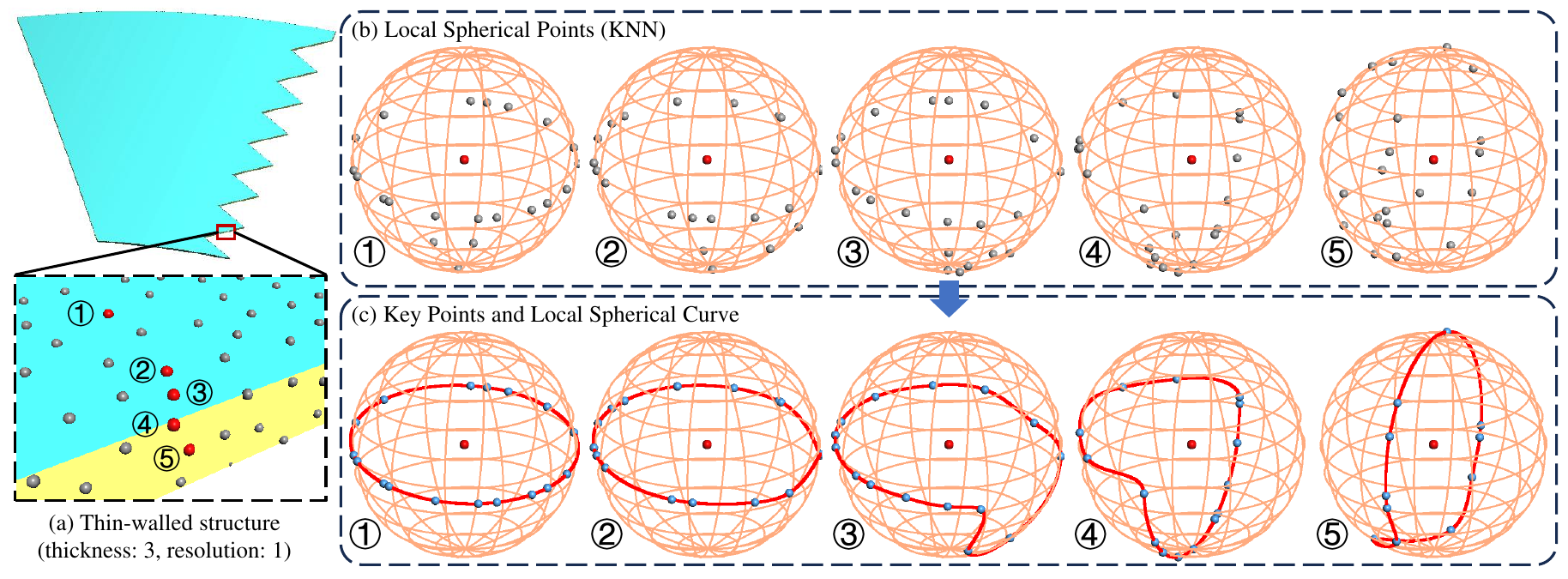}
\caption{Neighborhood distribution of five surface points on the unit sphere. Local spherical points are shown in grey, sampled key points on the fitted spherical Curve in blue, and local spherical curves in red. We observe that the neighborhoods of edge points resemble two semicircles (e.g. \textcircled{4}), while those of non-edge points maintain a complete great circle (e.g. \textcircled{1}).}
\label{fig:LSC}
\end{figure*}

The proposed method extracts edges from thin-walled structures, as illustrated in \cref{fig:overview}, through three main steps: local spherical curve construction, edge point identification, and edge optimization.
First, for each point, we project its sparse neighborhood onto a unit sphere, constructing a closed spherical curve to represent the local neighborhood (Sec~\ref{sec:LSC}).
Next, we use spherical harmonics (SH) to encode this spherical curve as a rotation-invariant local descriptor, which is then passed through an MLP layer to classify the target point as an edge or non-edge point (Sec~\ref{sec:extract}).
Finally, we leverage this descriptor again to estimate the normal vector, allowing us to formulate a projection function that refines the edge point positions (Sec~\ref{sec:optimize}).

\subsection{Local Spherical Curve Construction}
\label{sec:LSC}
Our key observation is that neighboring points co-planar with a target point align along great circle arcs after spherical projection, allowing the overall shape to be approximated as a closed, continuous curve.
Thus, given a point cloud $P = \{ p_i \in \mathbb{R}^3 | i = 1, 2, \ldots, N \}$, we first perform spherical normalization on each point $p_i$'s neighborhood point to obtain the local spherical point set $Q_i$, and then fit a closed, continuous curve.

Specifically, by taking $p_i$ as the center of the sphere, we normalize the radius of its neighboring points to 1, projecting them onto the unit sphere.
The normalization process is defined as follows:
\begin{equation}
Q_i = \{ q_j = \frac{p_j - p_i}{\| p_j - p_i \|} \Big| \, p_j \in \operatorname{\textit{KNN}}(p_i) \},
\label{Eq:normalization}
\end{equation}
where $q_j$ represents the local spherical point and $p_j$ is the \textit{KNN} neighboring points of $p_i$. Next, we apply Principal Component Analysis (PCA) to $Q_i$, projecting it onto a local plane. We then compute the convex hull of the projected points and select the vertices of this convex hull as the key point set $K_i$ for curve fitting:
\begin{equation}
K_i = \text{ConvexHull}\left(\text{PCA}(Q_i)\right).
\label{Eq2}
\end{equation}
Finally, we perform spline fitting on $K_i$ to obtain a spherical curve, effectively representing the local neighborhood.

\cref{fig:LSC} shows the distribution of local spherical points (grey), key points (blue), and the local spherical curve (red) for five target points across a thin-walled structure. Points \textcircled{1} and \textcircled{2}, both located on the top surface, their neighborhoods are close to a complete great circle despite interference from points on other surfaces. Point \textcircled{5}, positioned on a side-end face with sparse sampling, is influenced by both upper and lower surfaces of the thin-walled structure; however, its local spherical curve still largely retains a great circle form. In contrast, points \textcircled{3} and \textcircled{4} are near a sharp edge, with \textcircled{4} closer to the edge than \textcircled{3}. The local spherical curves of \textcircled{3} and \textcircled{4} are roughly formed by two semicircles, presenting a different pattern compared to points \textcircled{1} and \textcircled{2}. These results demonstrate that the proposed local spherical curve can robustly and sensitively capture edge features.

\subsection{Edge Point Identification}
\label{sec:extract}
After obtaining the spherical curve for each point, we use SH, an effective time-frequency domain analysis technique on a sphere, to construct rotation-invariant descriptors. This SH-encoded curve serves as a rotation-invariant local descriptor, which is then fed into an MLP to predict whether each point is an edge point.

In detail, we discretize the local spherical curve into a set of scattered points.
Each points $(x,y,z)$ is mapped to spherical coordinates $f(\theta,\phi)$ on the unit sphere, where the mapping is defined as $(\theta,\phi)=(\arccos{z},\arctan{y/x})$.
Here, $\theta$ and $\phi$ represent the zenith and azimuth angles, respectively.
Let $f_1, f_2, ..., f_n$ be a series of mapping on spherical space. 
We employ kernel density estimation~\cite{davis2011remarks} to construct a smooth, continuous function based on these sample points:
\begin{equation}
\label{eq:f}
f(\theta, \phi)=\frac{1}{n h} \sum_{i=1}^n k\left(\frac{\delta_i(\theta, \phi)}{h}\right),
\end{equation}
where $k$ is a kernel function, defined as a normal probability density function:
\begin{equation}
\label{eq:K}
k(x)=1/ \sqrt{2\pi} \exp \left(-0.5 x^2\right).
\end{equation}
The term $\delta_i(\theta, \phi)$ denotes the geodesic distance between $f_i(\theta_i, \phi_i)$ and $(\theta, \phi)$.
The parameter $h$, known as the window width or smoothing parameter, modulates the influence of mapping points $f_i$ to the density estimation at $(\theta, \phi)$.
It is defined as $h = \pi/B$, where $B$ represents the spherical function bandwidth.

To compute the SH coefficients for the band-limited function, we apply the discrete SH transform. 
Following the Shannon-Nyquist sampling theorem~\cite{driscoll1994computing}, we sample the model equiangularly at twice the bandwidth $B$ in both latitude and longitude, resulting in $2B\times2B$  sampling points $\left({{\theta }_{j}},{{\phi }_{k}} \right)\in \left( 0,\pi  \right)\times \left[ 0,2\pi  \right)$, with
\begin{equation}
\label{eq:grid}
    \left\{
    \begin{aligned}
        \theta_j&=\pi(2j+1)/4B&, \,  j=0,1, \ldots,2B-1 \\
        \phi_k&=\pi k/B&,\,  k=0,1, \ldots,2B-1
    \end{aligned}
    \right.
\end{equation}

The SH coefficients of the function $f(\theta,\phi)$ are obtained through the discrete SH transform:
\begin{equation}
\label{eq:DSHT}
\hat{f}_{l}^{m}=\frac{\sqrt{2\pi}}{2B}\sum_{j=0}^{2B-1}\sum_{k=0}^{2B-1}{{w}_{j}}f({{\theta}_{j}},{{\phi}_{k}}){{e}^{-im{{\phi}_{k}}}}P_{l}^{m}(\cos{{\theta}_{j}})
\end{equation}
for degree $0<l<B$ and order $\left| m \right|\le l$. 
Here $P_{l}^{m}$ denotes the associated Legendre functions defined on $\left[ -1,1 \right]$. 
The weights $w_j$ are computed using the method outlined in~\cite{driscoll1994computing}.

The SH bases are indexed by two integers, $l$ and $m$, where $-l\le m\le l$, resulting in $2l+1$ coefficients for each degree $l$. 
The parameter $l$ determines the frequency of the basis~\cite{kazhdan2003rotation}.
Consequently, the $2l+1$ coefficients corresponding to the same degree $l$ are combined into a vector:
\begin{equation}
\label{eq:degree_vector}
{\beta}_{l}=\{\hat{f}_{l}^{m}:m=-l,\ldots,0,\ldots,l \},
\end{equation}
and the $\ell_2$ norm $\|{\beta}_{l}\|$ is taken as the energy of the function $f$ in degree $l$. 
At last, the rotation-invariant local descriptor $D$ of function $f(\theta,\phi)$ with all $B$ degrees is defined as
\begin{equation}\label{eq:SD}
D(f,B)=\{\|{{\beta}_{l}\|}: l=0,\ldots,{B-1}\}.
\end{equation}  

After computing this descriptor, we calculate the average descriptor for ground-truth edge points and measure the feature difference between each point and this average descriptor. By applying a threshold to this difference, we can determine whether a point is classified as an edge point.
However, in our initial testing, we found that the threshold-based method struggles to effectively handle different edge styles. To address this, we use an MLP to adaptively learn from the rotation-invariant local descriptor, enabling accurate edge point identification.
Detailed network structure is provided in the supplementary material.

\begin{figure}[t]
\centering
\includegraphics[width=0.9\linewidth]{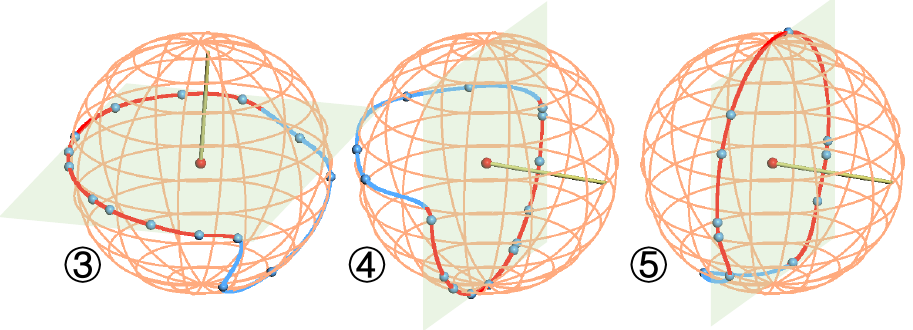}
\caption{Normal estimation results for points \textcircled{3}, \textcircled{4}, and \textcircled{5} in \cref{fig:LSC} (a). In each sub-figure, the red curve is the great circle arc extracted by RANSAC, the blue curve is the non-great circle segment, and the yellow arrow is the estimated normal vector.}
\label{fig:normal}
\end{figure}

\subsection{Edge Point Optimization}
\label{sec:optimize}
We observe that the initially identified edge points are roughly scattered around the true edge. While these points are linearly distributed overall, they form a band-like structure cross two intersecting surfaces when viewed locally. To improve accuracy, we propose constraining edge points to surface intersection regions. This approach is motivated by the fact that an edge point at the intersection of two surfaces has a zero projection distance to both surfaces.

Let $p_i$ represent an initially-identified edge point. 
Inspired by \cite{xu2022rfeps} and \cite{chen2024thin}, we introduce the following optimization function:
\begin{equation}                
\min_{z_i} \sum_{p_j \in \operatorname{\textit{KNN}}(p_i)}((z_i-p_j) \cdot n_j)^2+\mu\left\|z_i-p_i\right\|^2,
\label{eq:optimize}
\end{equation}
where $n_j$ is the normal vector of $p_j$, and $\mu$ is a balance parameter. The first term encourages $z_i$ to stay close to the tangent planes of neighboring points, while the second term penalizes the deviation of $z_i$ from its initial position $p_i$. In all our experiments, $\mu$ is set to a default value of 0.1.

\noindent\textbf{Normal estimation.}
Note that the edge point optimization method described above relies on accurate normal vectors; however, conventional normal estimation methods is unreliable in noisy or sparse neighborhood conditions. To resolve it, we propose a normal estimation algorithm based on the proposed local spherical curves, which can enhance robustness due to its underlying structure awareness.

As illustrated in \cref{fig:normal}, we observe that points co-planar with $p_i$ (shown on the red curve) tend to align on the same plane as the great circle arc after projection onto a sphere. In contrast, other points (shown on the blue curve) lie on a different plane.
Therefore, we use this great circle arc (red) on the local spherical curve to approximate the local plane and derive the normal vector.
Specifically, we employ random sampling consensus (RANSAC) to identify the great circle arc on the local spherical curve to determine the normal plane, and use the rest (blue) to determine the normal orientation.
\cref{fig:normal} shows the normal estimation of the neighborhood \textcircled{3}, \textcircled{4}, and \textcircled{5} in \cref{fig:LSC} (a). 

\begin{table*}[h]
    \setlength{\tabcolsep}{8pt}
    \centering
    \small
    \caption{Quantitative comparison of edge point extraction on the thin-walled structure dataset. `-' denotes a failed calculation.}
    \begin{tabular}{ccccccc} 
        \hline
        &EC-Net~\cite{yu2018ec} &PIE-Net~\cite{wang2020pie}  &MFLE~\cite{chen2021multiscale} &RFEPS~\cite{xu2022rfeps} & STAR-Edge (Ours)\\
        \hline
        ECD$\downarrow$    & 0.2936 & 52.1204 & 0.4839 & 1.4392 & \textbf{0.0587} \\
        Recall$\uparrow$   & -      & 0.0503  & 0.9755 & \textbf{0.9997} & 0.9952 \\
        Precision$\uparrow$   & - & 0.0295 & 0.3028 & 0.1512 & \textbf{0.5418} \\
        F1$\uparrow$      & - & 0.0365 & 0.4621 & 0.2625 & \textbf{0.7013} \\
        Accuracy$\uparrow$     & - & 0.9349  & 0.9470 & 0.9544 & \textbf{0.9931} \\
        \hline
    \end{tabular}
    \label{tab:sota}
\end{table*}

\section{Experiments and Discussion}
\label{sec:Experiment And Analysis}
In this section, we evaluate the effectiveness of our method by comparing it to several learning-based approaches~\cite{yu2018ec, wang2020pie, zhu2023nerve} and two traditional methods~\cite{xu2022rfeps, chen2021multiscale}. We then assess its robustness under varying noise intensities and point cloud resolutions and conduct an ablation study to examine the impact of specific design choices.

\subsection{Experiment Setup}
\noindent\textbf{Dataset.}
Given the characteristics of thin-walled structures (small dimensions in the thickness direction), we selected a dataset~\cite{chen2024thin} specifically focused on thin-walled structures for our experiments. This dataset includes 42 distinct 3D models with curved edges of varying shapes, sizes, and curvatures, and thicknesses. For a more detailed description of this dataset, please refer to the supplemental file. Note that a subset of 32 shapes is used for training, while 10 shapes are reserved for testing. For each shape, we sample point clouds at resolutions of 0.3, 0.5, and 0.8. Additionally, three levels of random Gaussian noise are added to the point clouds at a resolution of 0.5.
Each sampled point cloud contains approximately 20,000 to 15 million points.

We also perform experiments on the well-known ABC dataset~\cite{koch2019abc}, which contains over one million CAD models with edge annotations. To evaluate the generalizability of our method, we select sharp-edged CAD models from the ABC dataset, following the setup in~\cite{zhu2023nerve}. This subset comprises 2,364 models, which are randomly divided into a training set (80\%) and a test set (20\%).
For each shape, 50,000 points are sampled.

\noindent\textbf{Implementation.}
All experiments are conducted on an Intel i7-11700 CPU and an NVIDIA RTX 4090 GPU. For constructing the local rotation-invariant descriptor, we use a \textit{KNN} query with $k=26$ and a spherical harmonics transform bandwidth of 10. 
The proposed approach is implemented in PyTorch and trained over 40 epochs using the Adam optimizer, with a batch size of 1024 and a learning rate of 0.02.

\noindent\textbf{Metrics.}
We quantitatively evaluate all methods across two aspects. First, point classification performance is assessed using Recall, Precision, F1-score, and Accuracy. 
Second, to evaluate the quality of the final edge point extraction, we introduce a new metric: Edge Chamfer Distance (ECD). This metric measures the similarity between the optimized edges and the ground-truth edges.
The introduction of ECD is necessary because the ground-truth sampled point cloud consists of dense points, while the optimized edges may vary in sparsity depending on the chosen resolution. Standard Chamfer Distance, without adjustment, could disproportionately penalize sparse edges, leading to inaccurate similarity assessments. ECD addresses this limitation by providing a more reliable comparison for edge extraction quality. 
Let $X$ represents the point cloud of the optimized edges, and $Y$ denote the ground-truth edges. 
The ECD is defined as follows:
\begin{equation}
\text{ECD}= \frac{1}{|X|} \sum_{p \in X} \min_{q \in Y} \| p - q \|_2^2
\label{eq:CD}
\end{equation}
Note that for each shape, we label sampled points within a specified distance threshold from the nearest edge as ground-truth edge points. We set this threshold to 0.3 for the dataset in~\cite{chen2024thin} and 0.025 for the ABC dataset.

\subsection{Evaluation on Thin-walled Dataset}

\begin{figure*}[!t]
\centering
\includegraphics[width=0.9\linewidth]{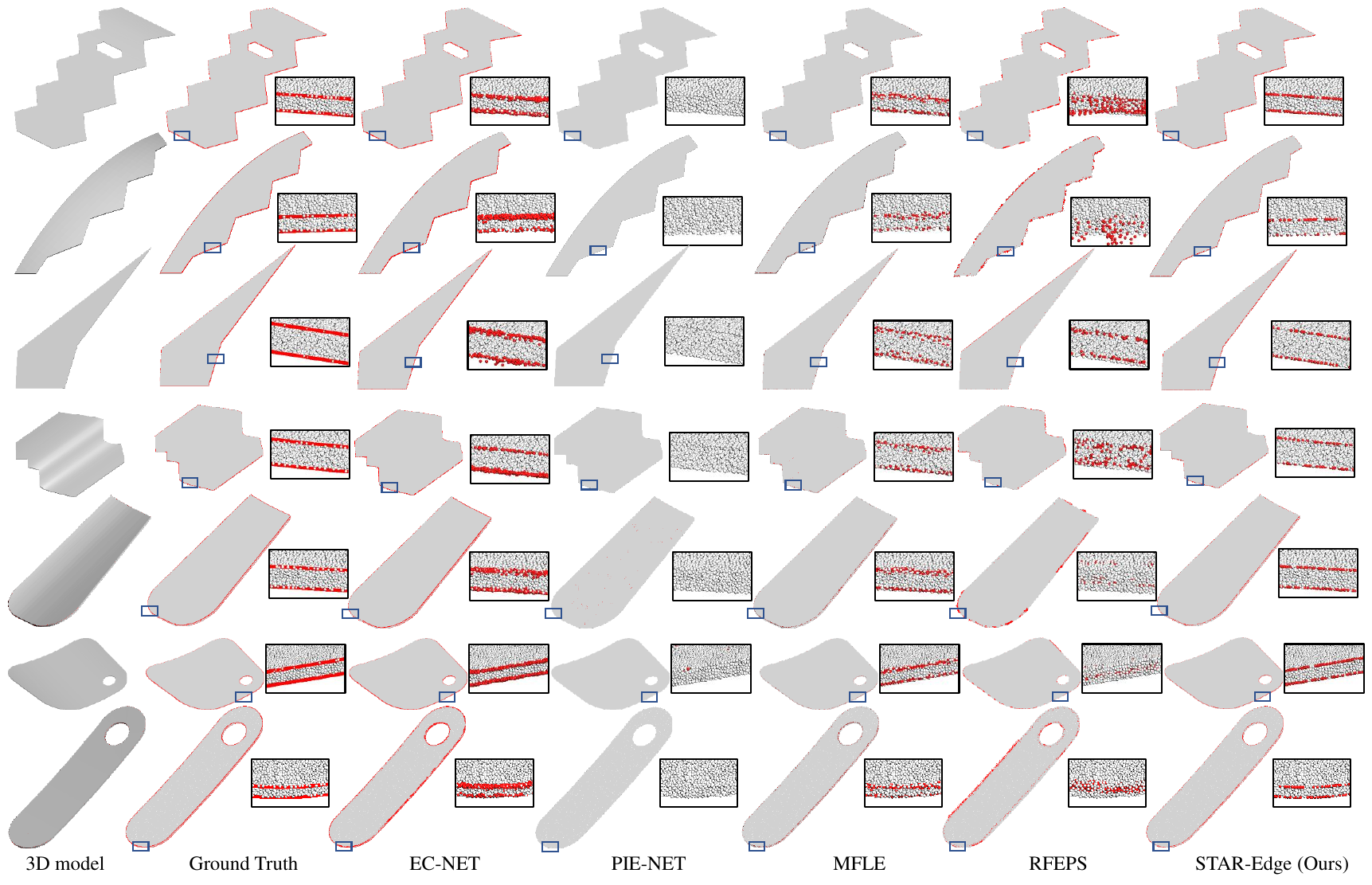}
\caption{Visual comparison of state-of-the-art methods on the thin-walled structure dataset. In these close-up views, our method demonstrates superior accuracy in edge point extraction. Edge points are in red color.}
\label{fig:sota}
\end{figure*}

\begin{figure}[t]
\centering
\includegraphics[width=0.9\linewidth]{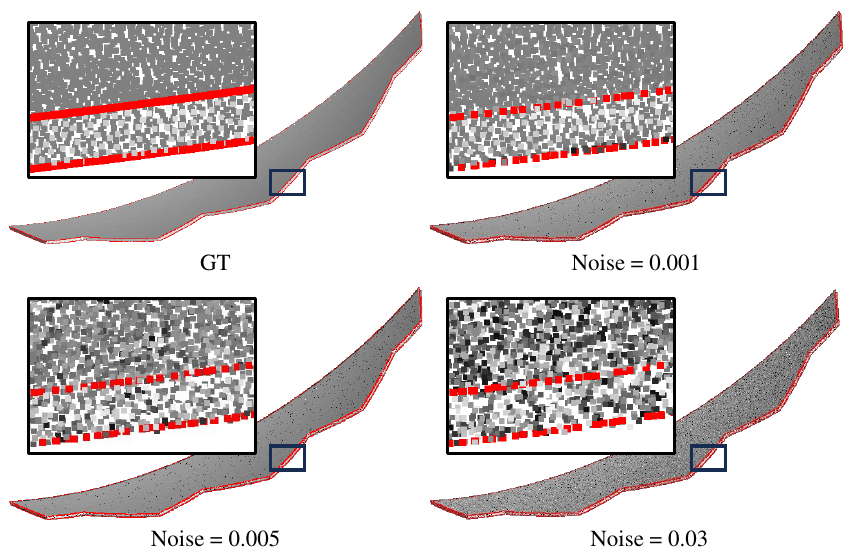}
\caption{Visual comparison under different noise levels.}
\label{fig:noise}
\end{figure}

\begin{figure}[t]
\centering
\includegraphics[width=0.9\linewidth]{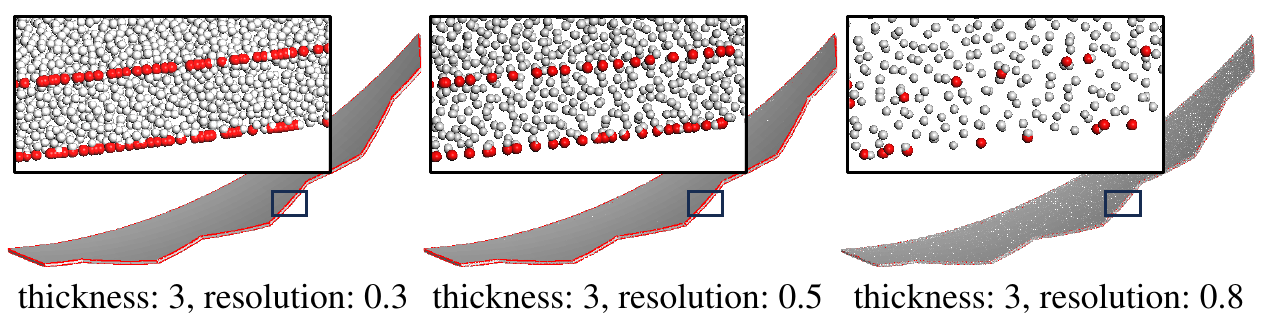}
\caption{Visual comparison under different resolutions.}
\label{fig:sample}
\end{figure}

In this section, we compare our STAR-Edge method with state-of-the-art approaches, including EC-Net~\cite{yu2018ec}, PIE-Net~\cite{wang2020pie}, NerVE~\cite{zhu2023nerve}, MFLE~\cite{chen2021multiscale}, and RFEPS~\cite{xu2022rfeps}, on the thin-walled dataset. All methods use point cloud data with a resolution of 0.5 as input.
To accommodate network size limitations, PIE-Net is provided with 8,096 uniformly sampled points obtained through farthest point sampling, while NerVE converts the point cloud into a voxel grid with a resolution of $64^3$. All networks are retrained on the same dataset for a fair comparison.

\cref{fig:sota} provides visual comparisons of all methods, while quantitative results are presented in \cref{tab:sota}.
Both results demonstrate that our approach outperforms others on thin-walled structures, achieving better accuracy and a more uniform edge point distribution.
Specifically, EC-Net, which up-samples point clouds in edge regions, tends to generate redundant points around true edges, leading to higher Edge Chamfer Distance (ECD). PIE-Net and NerVE perform less effectively on this dataset; PIE-Net requires dense sampling to ensure its effectiveness in edge extraction, while NerVE’s voxel size exceeds the thin-wall thickness, making it struggle to perceive edges. A potential improvement for these methods is to segment small patches for re-processing, as discussed in the supplemental file.
The geometry-based methods, MFLE and RFEPS, also show limitations: MFLE requires frequent parameter adjustments, while RFEPS produces noisier edge results.
Contrast, our method leverages the local spherical curve to achieve more precise edge points and displacement vectors, resulting in superior performance for edge detection in thin-walled structures. 

\noindent\textbf{Running time.} The running times of various methods on point clouds containing approximately 320,000 points are as follows: our method required 214.0 seconds, MFLE completed in 1.14 seconds, RFEPS took 19 minutes, ECNet finished in 83 seconds, and PIE-Net was the fastest, consuming less than 1.0 second. We attribute PIE-Net's exceptional speed to its down-sampling process. More detailed statistics can be found in the supplementary materials.

\subsection{Robustness Study}
We conducted robustness testing for the proposed STAR-Edge by introducing varying levels of noise and resolution. In our experiments, the thin-walled shapes have a thickness of 2–3 (which is very small relative to other shape dimension). At a resolution of 0.3, approximately 9 points can be sampled on the side-end surface, whereas at a resolution of 0.8, it becomes challenging to obtain even 4 uniformly distributed points.
Testing at a resolution of 0.8 poses a significant robustness challenge, as the limited number of sampling points reduces the reliability of projections onto spherical curves.

\cref{tab:robustness1} and \cref{tab:robustness2} present the effects of noise intensity and resolution on the predictions of our STAR-Edge method. We observe that, even under high noise intensity or significantly reduced resolution, our method consistently produces reasonable edge point predictions with low Edge Chamfer Distance (ECD).
As shown in \cref{fig:noise}, when the noise level reaches 0.003, the detection performance remains largely unaffected.
As shown in \cref{fig:sample}, demonstrates that even in an extreme case with a thickness of 3 and a resolution of 0.8, the upper and lower edges can still be clearly distinguished.
Additionally, in \cref{tab:robustness2}, we compare the performance of STAR-Edge with state-of-the-art methods across inputs with varying noise levels and resolutions.
Our method achieves the best numerical performance.

\begin{table}[t]
    \setlength{\tabcolsep}{4pt}
    \centering
    \small
    \caption{Quantitative classification results under different noise levels and resolutions.} 
    \begin{tabular}{c|c c c|c c c} 
        \hline
         & \multicolumn{3}{c|}{Noise} & \multicolumn{3}{c}{Resolution} \\
        
                    &  $0.001$ & $0.005$ & $0.03$ & $0.3$ & $0.5$ & $0.8$ \\
        \hline
        ECD$\downarrow$     & 0.059 & 0.059 & 0.199 & 0.048 & 0.059 & 0.261 \\
        Recall$\uparrow$    & 0.995 & 0.995 & 0.994 & 0.960 & 0.995 & 0.988 \\
        Precision$\uparrow$ & 0.542 & 0.541 & 0.523 & 0.784 & 0.542 & 0.582 \\
        F1$\uparrow$        & 0.701 & 0.701 & 0.685 & 0.863 & 0.701 & 0.731 \\
        Accuracy$\uparrow$  & 0.993 & 0.993 & 0.993 & 0.998 & 0.993 & 0.993 \\
        \hline
    \end{tabular}
    \label{tab:robustness1}
\end{table}

\begin{table}[t]
\setlength{\tabcolsep}{2pt}
    \centering
    \small
    \caption{Quantitative ECD comparisons under different noise and resolution. `-' indicates that the calculation failed.}
    \begin{tabular}{l|c c c|c c c} 
        \hline
        \multirow{2}{*}{Methods}& \multicolumn{3}{c|}{Noise} & \multicolumn{3}{c}{Resolution} \\
        &  $0.001$ & $0.005$ & $0.03$ &0.3& $0.5$ & $0.8$ \\
        \hline
        EC-Net  & 0.295 & 0.293 & 0.294 & - &0.294 & 0.518 \\

        PIE-Net  & 50.285 & 52.332 & 49.156 & 50.791 & 52.120 & 52.222 \\

        RFEPS    & 1.440 & 1.436 & 1.442 & 0.391 & 1.439 & 10.341 \\

        MFLE     & 0.484 & 0.453 & 0.409 & 0.406 & 0.484 & 0.742 \\

        Ours    & \textbf{0.059} & \textbf{0.059} & \textbf{0.199} & \textbf{0.048} & \textbf{0.059} & \textbf{0.260} \\
        \hline
    \end{tabular}
    \label{tab:robustness2}
\end{table}

\subsection{Ablation Study} 

\cref{tab:ablation} reports the performance of our method under varying SH bandwidths, which determine the precision and detail of the local feature representation. The results indicate that performance improves with increasing bandwidth; however, higher bandwidth settings also result in increased computational cost.
Notably, the performance at a bandwidth of 10 is comparable to that at 15, making a bandwidth of 10 the optimal choice for balancing accuracy and efficiency.

\subsection{Evaluation on ABC dataset}
\begin{figure}[t]
\centering
\includegraphics[width=0.95\linewidth]{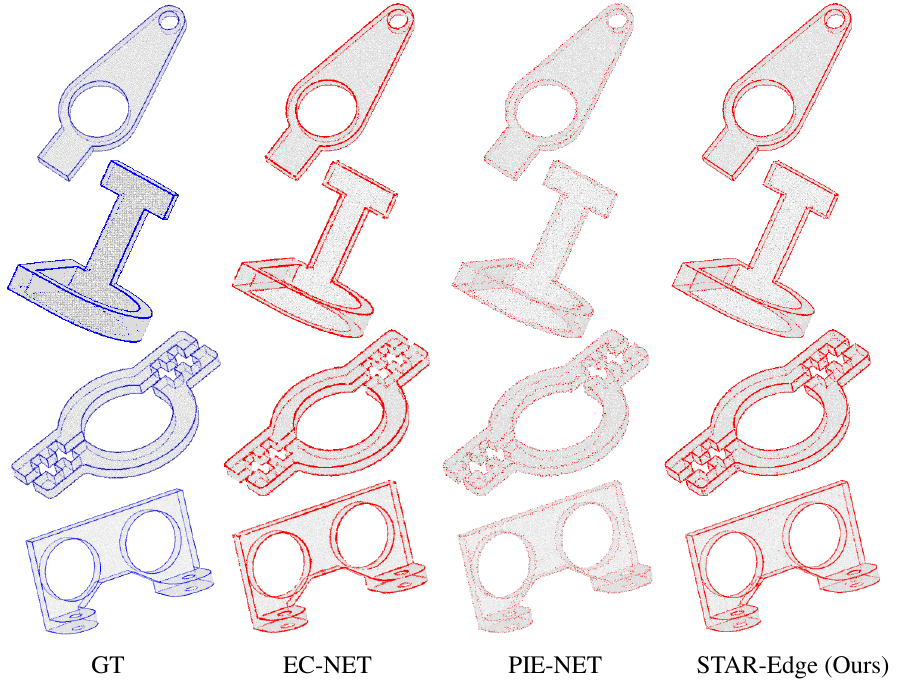}
\caption{Visual comparison of edge extraction results on 3D shapes from the ABC dataset. Our method achieves more precise edge extraction, particularly in the sharp corner regions.}
\label{fig:abc}
\end{figure}

We evaluate our method on the ABC dataset following the experimental setup in~\cite{zhu2023nerve}.
Quantitative results are presented in \cref{tab:abc}, which demonstrate that STAR-Edge consistently outperforms all baseline methods in terms of edge extraction accuracy. 
Visual comparisons in \cref{fig:abc} further highlight the superior performance of our method, as it achieves more precise edge extraction, particularly in the sharp corner regions.


\begin{table}[t]
    \setlength{\tabcolsep}{2pt}
    \centering
    \small
    \caption{Ablation study about SH bandwidths.} 
    \begin{tabular}{c|c c c c c } 
        \hline
        Bandwidth  &  Recall$\uparrow$ & Precision$\uparrow$ & F1$\uparrow$ & Accuracy$\uparrow$ & ECD$\downarrow$  \\
        \hline
        5       & 0.988 & 0.523 & 0.685 & 0.991 & 0.122  \\
        10       & 0.995 & 0.542 & 0.701 & 0.993 & 0.059  \\
        15       & 0.995 & 0.583 & 0.785 & 0.998 & 0.048  \\
        \hline
    \end{tabular}
    \label{tab:ablation}
\end{table}

\begin{table}[ht]
    \setlength{\tabcolsep}{5pt}
    \centering
    \small
    \caption{Quantitative comparison on the ABC dataset.} 
    \begin{tabular}{c|cccccc} 
        \hline
        &EC-NET & PIE-NET & MFLE & RFEPS & Ours\\
        \hline
        ECD$\downarrow$    & 0.0037 & 0.0090 & 0.0049 & 0.0026 & \textbf{0.0024} \\
        CD$\downarrow$    & 0.0226 & 0.0037 & 0.0074 & 0.0012 & \textbf{0.0074 }\\
        \hline
    \end{tabular}
    \label{tab:abc}
\end{table}

\section{Conclusion}
In this work, we presented STAR-Edge, a novel point cloud edge extraction approach specifically designed for thin-walled structures. The core of our method is the introduction of a new structure-aware neighborhood representation—the local spherical curve. This representation effectively emphasizes co-planar points while minimizing interference from unrelated surfaces, enabling robust and accurate edge detection in complex thin-walled geometries.




{
    \small
    \bibliographystyle{ieeenat_fullname}
    \bibliography{main}
}

\clearpage
\setcounter{page}{1}
\maketitlesupplementary

\section{Appendix}
This appendix provides details on network architecture for edge detection, the visualization and composition of the thin-walled structure dataset, and the evaluation of PIE-NET with patch-based improvements.
We also show more comparison visualization results of STAR-Edge, and we also verify its effectiveness with real scanned point clouds.
We then perform ablation studies on normal estimation and edge optimization as well as a running time analysis.
Finally, the limitations of the proposed approach are discussed.

\subsection{Network architecture for point classification}

\begin{figure}[ht]
\centering
\includegraphics[width=1\linewidth]{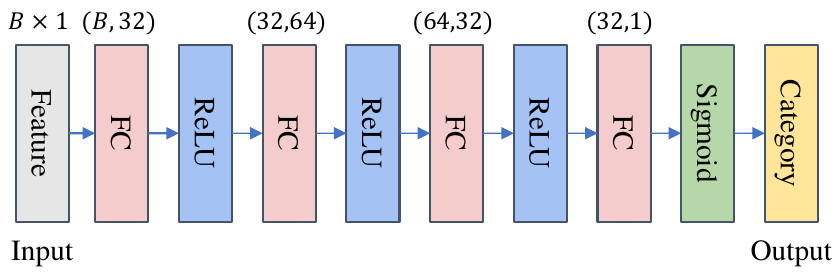}
\caption{Detailed network architecture for edge point classification.}
\label{fig:structure}
\end{figure}

\cref{fig:structure} shows the used classification network composed of a series of fully connected (FC) layers. The input layer accepts features of dimension $B$, representing the bandwidth of the spherical harmonics. The output layer produces a categorical label, predicting whether a given point is an edge point or not.

\subsection{Thin-walled structure dataset}
\begin{figure}[ht]
\centering
\includegraphics[width=1\linewidth]{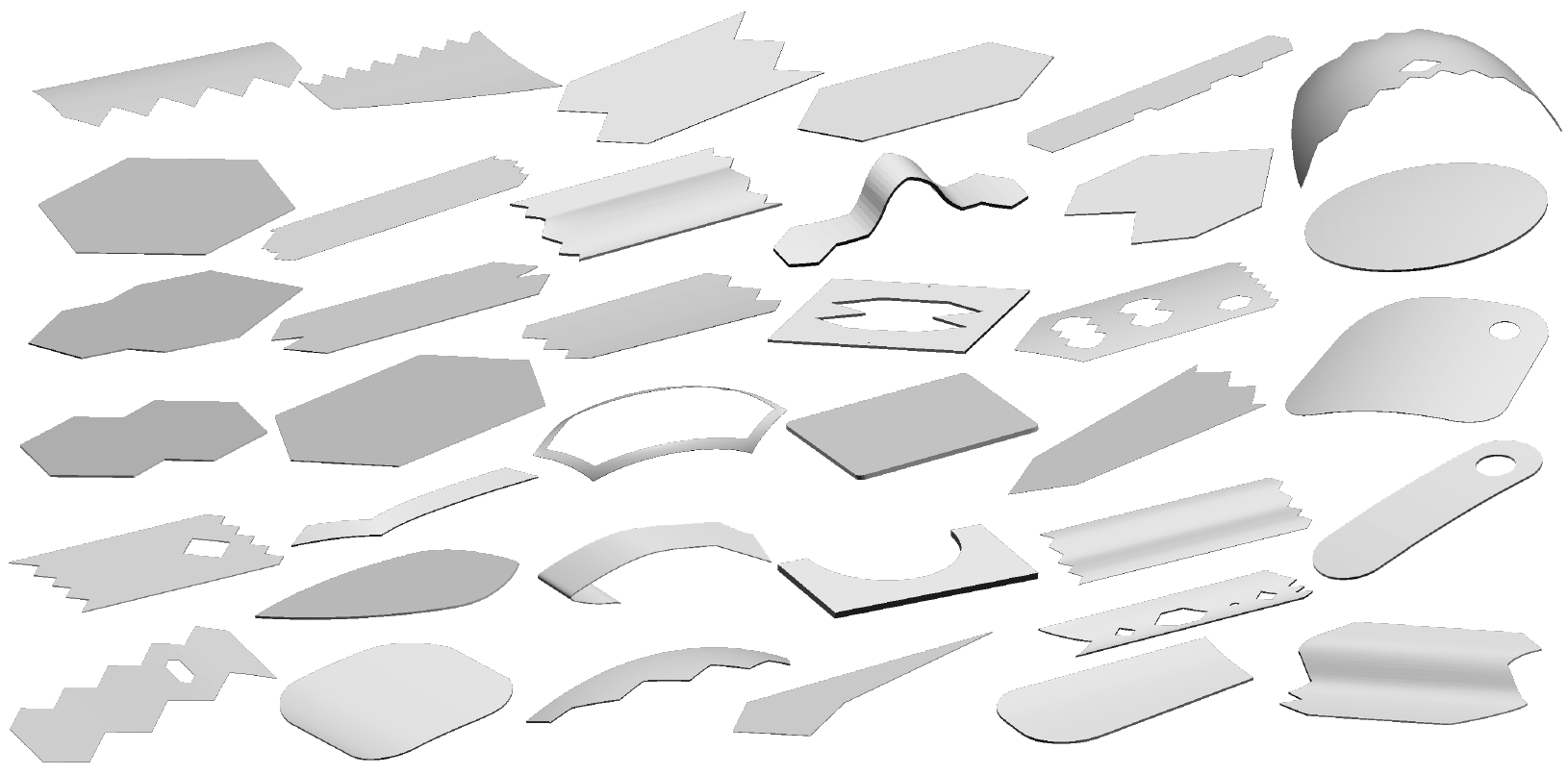}
\caption{Visualization of thin-walled structure dataset.}
\label{fig:thin-walled}
\end{figure}

\cref{fig:thin-walled} visualizes the used thin-walled structure dataset. These structures exhibit diverse curved edges with varying shapes, sizes, curvatures, and thicknesses ranging from 2 to 4. 
The dataset comprises 42 unique 3D models, of which 32 are used for the training set and 10 for the test set.

\subsection{More careful comparison with PIE-Net}
\label{sec:supp_pie}

To compare with the SOTA method, PIE-Net, in our early tests, we applied Farthest Point Sampling (FPS) to thin-walled shapes and fed the downsampled point clouds into PIE-Net to extract edges. However, as shown in the main paper, PIE-Net performed poorly on the thin-walled structure dataset. Upon analysis, we determined that the primary issue was the sparsity of the downsampled version of the input data.
To address this, we normalized the point clouds, divided each shape into patches, and processed each patch individually through PIE-Net. The extracted edge points from all patches were then aggregated to reconstruct the full edge. The results of both implementations are presented in \cref{tab:pienet}. While the patch-based method showed slightly better results on the ECD metric, its performance remained sub-optimal compared to other state-of-the-art methods.
We attribute this limitation to the presence of nearby surfaces along the edges of thin-walled shapes. The \textit{KNN}-based feature extraction approach employed by PIE-Net struggles to accurately capture and represent this specific characteristic.

\begin{table}[ht]
    \setlength{\tabcolsep}{4pt}
    \centering
    \small
    \caption{Comparisons of different PIE-NET implementations.}
    \begin{tabular}{c|c|c}
    \hline
        Metric & PIE-Net (fps)~\cite{wang2020pie} & PIE-Net (patch)~\cite{wang2020pie} \\
    \hline
        Recall$\uparrow$    & 0.05  & \textbf{0.275} \\
        Precision$\uparrow$ & \textbf{0.03}  & 0.019 \\
        F1$\uparrow$        & \textbf{0.037} & 0.035 \\
        Accuracy$\uparrow$  & \textbf{0.935} & 0.829 \\
        ECD$\downarrow$     & 52.12 & \textbf{43.92} \\
    \hline
    \end{tabular}
    \label{tab:pienet}
\end{table}

\subsection{More visualization results on thin-walled structures and ABC dataset}
\begin{figure*}[ht]
\centering
\includegraphics[width=1\linewidth]{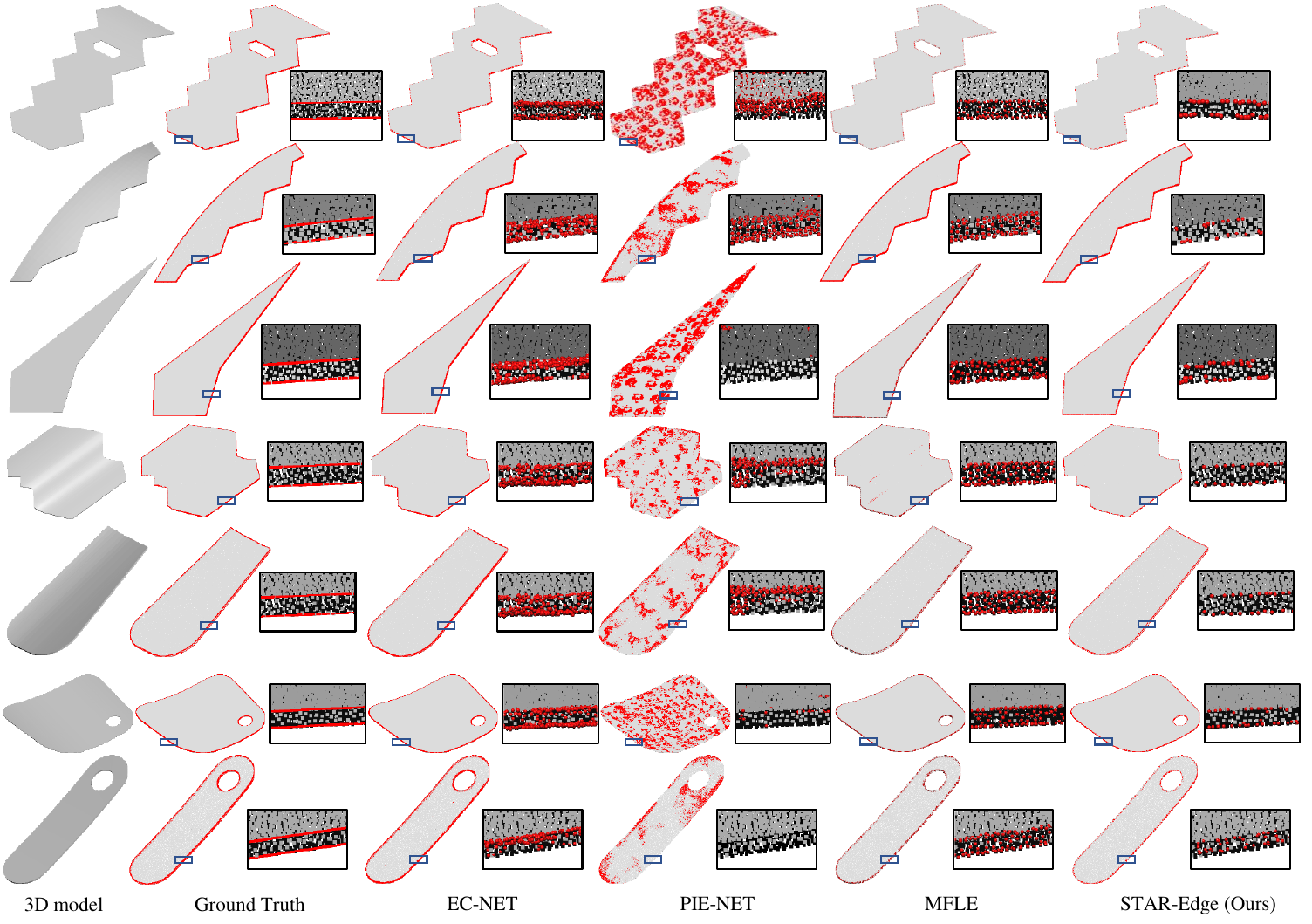}
\caption{Visual comparison of different methods on the thin-walled structure dataset at a resolution of 0.8. Red points represent edge points. Our method exhibits superior accuracy in extracting edge points.} 
\label{fig:resolution_0.8}
\end{figure*}

\cref{fig:resolution_0.8} compares the performance of STAR-Edge with state-of-the-art methods on the thin-walled structures under a challenging sampling resolution. Specifically, the tested thin-walled shapes have thicknesses ranging from 2 to 3, with a sampling resolution of 0.8. This implies that side-ended faces may be covered by as few as three points.

In these scenarios, EC-Net and MFLE tend to generate redundant points along the ground-truth edges, including a significant number of misclassified points. As analyzed in Sec.~\ref{sec:supp_pie}, PIE-Net also performs poorly on the thin-walled structure dataset. In contrast, our method outperforms these baselines by producing more precise and clearer edge points.

\begin{figure*}[ht]
\centering
\includegraphics[width=1\linewidth]{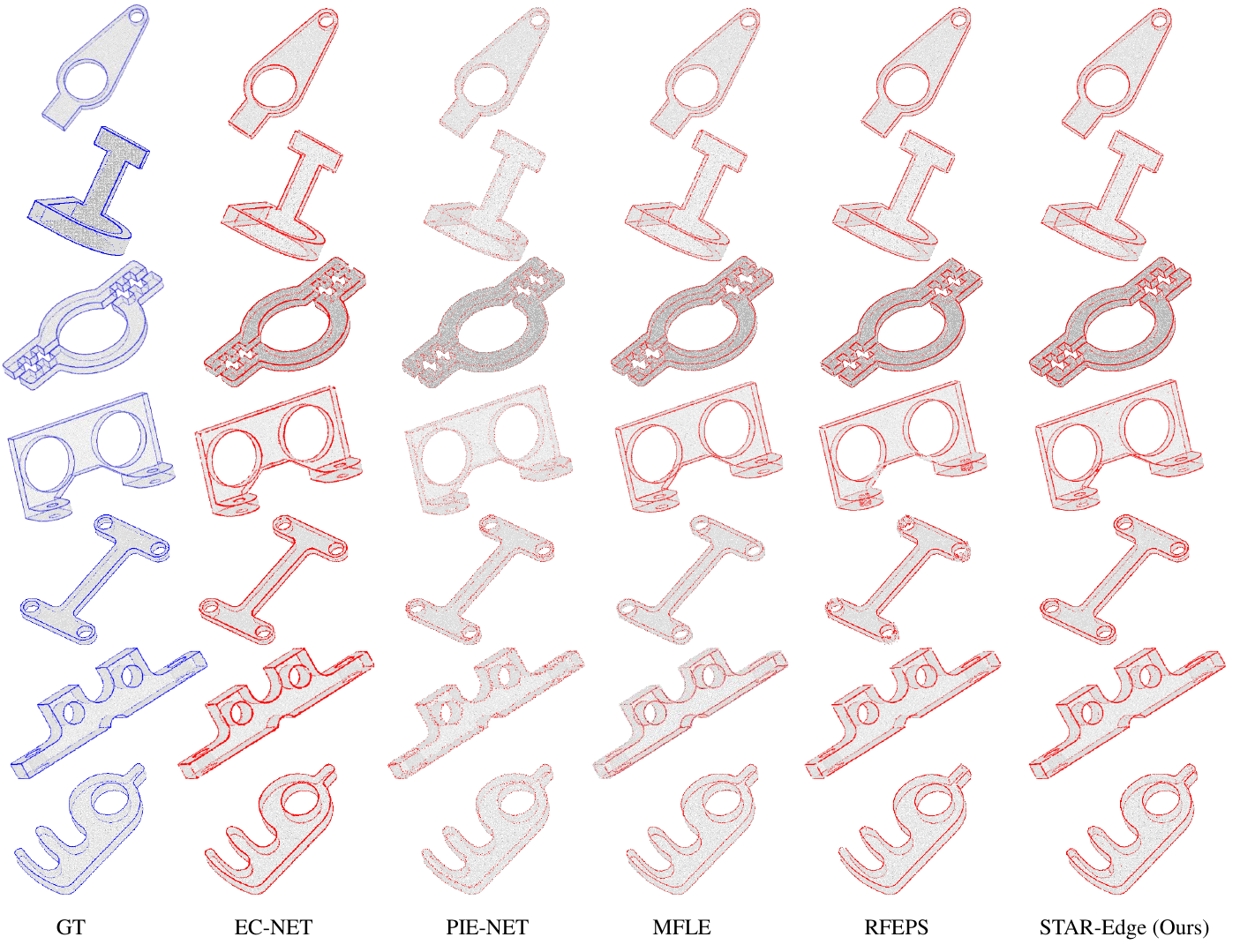}
\caption{Visual comparison of edge extraction results on 3D shapes from the ABC dataset.} 
\label{fig:abc_supp}
\end{figure*}

We also evaluate our method on the ABC dataset. As shown in \cref{fig:abc_supp}, STAR-Edge achieves more precise edge extraction than other baselines. It effectively captures critical features, particularly in regions with sharp corners.

\subsection{Evaluation on Real-scanned 3D point clouds}
\begin{figure*}[ht]
\centering
\includegraphics[width=1\linewidth]{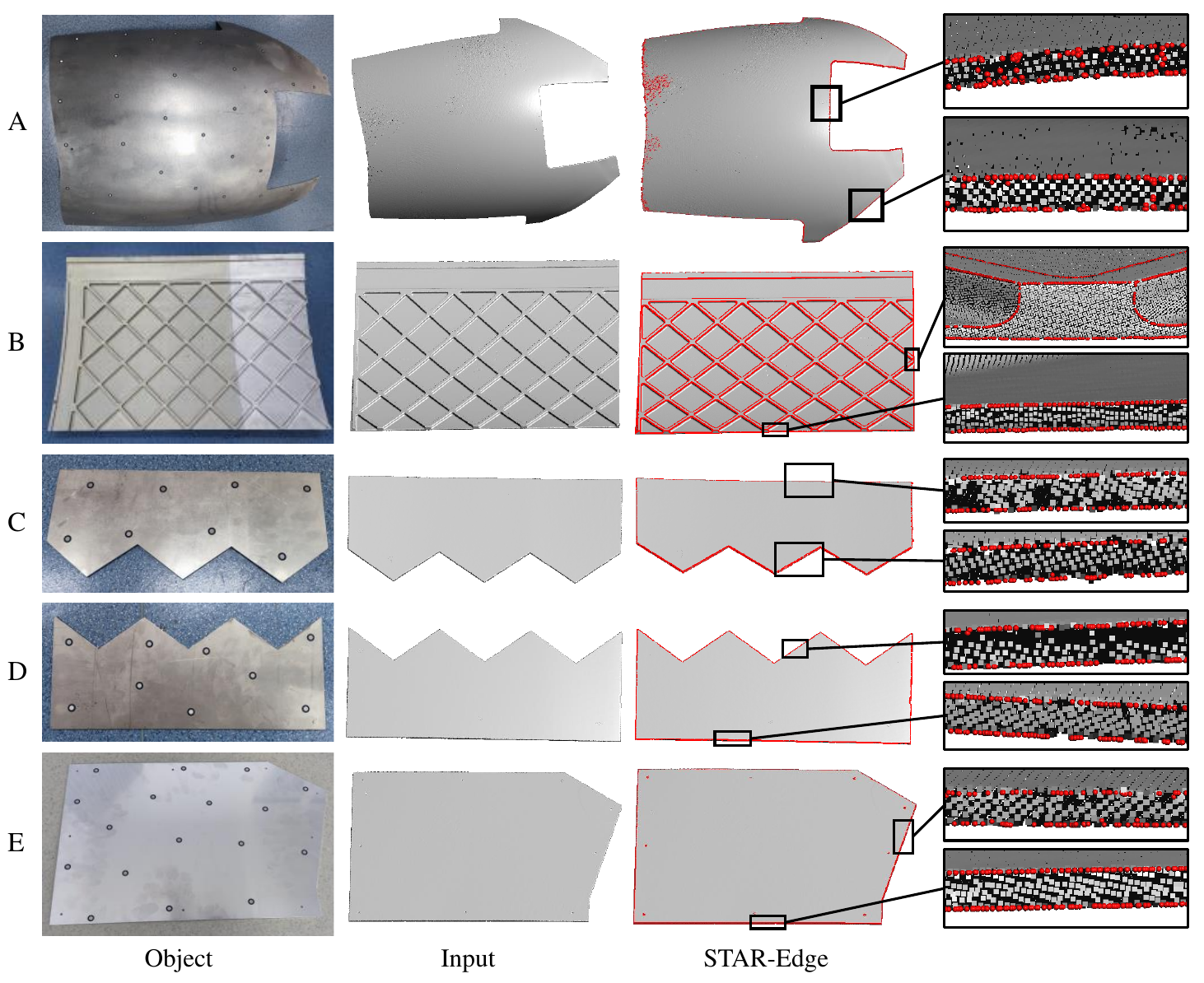}
\caption{Visualization of our method applied to real-scanned thin-walled structure data. In these close-up views, our method effectively extracts edge points with high accuracy. Edge points are colored in red.} 
\label{fig:real}
\end{figure*}

In addition to the synthetic data, we evaluate our method on several real-scanned point clouds of practical thin-walled workpieces. Object A is a 1-meter-long skin panel with a thickness of 2 mm. Object B is a wall panel with a minimum thickness of approximately 2.5mm. Objects C, D, and E are thin-walled structures with a uniform thickness of 3mm. All point cloud data were collected using the SIMSCAN scanner, configured with a resolution of 0.5mm.

\cref{fig:real} presents the edge extraction results on these scans. Notably, Object A poses a significant challenge due to its extremely thin-walled structure, where the cross-sectional side edges are represented by only about four points. Despite some misidentifications, STAR-Edge effectively identifies edge points and demonstrates robustness against interference from the very close upper and lower surfaces.

\subsection{Effect of different normals for edge point optimization}

To evaluate the impact of different normal estimation methods on the final edge point results, we compare two variants of STAR-Edge, as detailed in \cref{tab:ablation_normal}.
\textit{Variant A}: The edge point optimization module is removed, relying solely on the classification results.
\textit{Variant B}: The edge point optimization uses the commonly employed PCA method for local normal estimation.
\textit{Ours}: The proposed normal estimation method is applied to optimize the edge points.
Both \textit{Variant A} and \textit{Variant B} exhibit noticeable performance degradation, highlighting the effectiveness of our normal estimation method in refining edge points.

\begin{table}[ht]
    \setlength{\tabcolsep}{8pt}
    \centering
    \small
    \caption{Effect of different normals for edge point optimization.} 
    \resizebox{0.5\textwidth}{!}{
    \begin{tabular}{c|c|c} 
    \hline
    Method Variants & Normal Estimation Method & ECD$\downarrow$ \\
    \hline
     \textit{A}: w/o optimization& -- &0.2921\\
     \textit{B}: w/ optimization& PCA normal & 0.1134\\
     \textit{Ours}: w/ optimization & ours normal & \textbf{0.0587}\\
    \hline
    \end{tabular}}
    \label{tab:ablation_normal}
\end{table}

\subsection{Running time performance}
\begin{table}[ht]
    \setlength{\tabcolsep}{6pt}
    \centering
    \small
    \caption{Running time for different number of points.} 
    \begin{tabular}{c|c|c|c} 
    \hline
    \#Points & 92,000 & 327,000 & 2,090,000 \\
    \hline
        EC-Net~\cite{yu2018ec} & 25s & 1min 23s & 11min 8s \\
        PIE-Net (fps)~\cite{wang2020pie} & 0.4s & 0.4s & 0.4s \\
        PIE-Net (patch)~\cite{wang2020pie} & 1.6s & 5.7s & 35.3s \\
        MFLE~\cite{chen2021multiscale} & 0.7s & 1.1s & 7.6s \\
        RFEPS~\cite{xu2022rfeps} & 6min 56s & 19min & 322min \\
        STAR-Edge (Ours) & 59.6s & 3min 30s & 22min 24s \\
    \hline
    \end{tabular}
    \label{tab:running time}
\end{table}

We report the running time statistics in \cref{tab:running time}, which include point clouds with different numbers of points. As reported, the running efficiency of our method is moderate.

\subsection{Limitations}
Our method demonstrates robust edge detection performance for thin-walled structure data.
However, it relies somewhat on the distribution of local neighborhood spherical projections, which can lead to misidentifications in the presence of sharp noise. 
Additionally, insufficient points on the side edges may hinder accurate detection. 
Due to the need for per-point neighborhood calculations and iterative optimization, our approach exhibits lower efficiency compared to other methods.

\end{document}